\documentclass[10pt,twocolumn,letterpaper]{article}

\usepackage{iccv}              

\definecolor{iccvblue}{rgb}{0.21,0.49,0.74}
\usepackage[pagebackref,breaklinks,colorlinks,allcolors=iccvblue]{hyperref}

\usepackage{amsmath,amsfonts}
\usepackage{algorithmic}
\usepackage{algorithm}
\usepackage{textcomp}
\usepackage{stfloats}
\usepackage{url}
\usepackage{verbatim}
\usepackage{mathrsfs}
\usepackage{mathtools}
\usepackage{multirow}
\usepackage{multicol}
\usepackage{glossaries}
\usepackage{graphicx}
\usepackage{array}
\usepackage{colortbl}
\usepackage{xcolor}
\usepackage{booktabs}
\usepackage{arydshln}
\usepackage{adjustbox}
\usepackage[capitalize]{cleveref}
\usepackage{pifont}
\usepackage{balance}

\definecolor{best}{rgb}{1.0, 0.6, 0.6}
\definecolor{second}{rgb}{0.98, 0.78, 0.57}
\definecolor{third}{rgb}{1.0, 1.0, 0.66}
\definecolor{cvprblue}{rgb}{0.21,0.49,0.74}
\def\BibTeX{{\rm B\kern-.05em{\sc i\kern-.025em b}\kern-.08em    T\kern-.1667em\lower.7ex\hbox{E}\kern-.125emX}}

\def\paperID{4} 
\def\confName{ICCV}
\def\confYear{2025}

\title{3D Gaussian Representations with Motion Trajectory Field for Dynamic Scene Reconstruction}
\author{Xuesong Li$^{1,2}$,  Lars Petersson$^{1}$, Vivien Rolland$^{1}$ \\
\small $^{1}$CSIRO \quad $^{2}$The Australian National University\\
{\tt\small xuesong.li@csiro.au}
}

\begin{document}

\maketitle
 \begin{abstract}

This paper addresses the challenge of novel-view synthesis and motion reconstruction of dynamic scenes from monocular video, which is critical for many robotic applications. Although Neural Radiance Fields (NeRF) and 3D Gaussian Splatting (3DGS) have demonstrated remarkable success in rendering static scenes, extending them to reconstruct dynamic scenes remains challenging. In this work, we introduce a novel approach that combines 3DGS with a motion trajectory field, enabling precise handling of complex object motions and achieving physically plausible motion trajectories. By decoupling dynamic objects from static background, our method compactly optimizes the motion trajectory field. The approach incorporates time-invariant motion coefficients and shared motion trajectory bases to capture intricate motion patterns while minimizing optimization complexity. Extensive experiments demonstrate that our approach achieves state-of-the-art results in both novel-view synthesis and motion trajectory recovery from monocular video, advancing the capabilities of dynamic scene reconstruction\footnote[1]{\url{https://benzlxs.github.io/3DGS_MTF/}}.

\end{abstract}
\section{Introduction}
The novel-view synthesis and motion reconstruction of dynamic scenes from a monocular video is crucial for various applications such as robot perception and virtual reality~\cite{li2023dynibar, tian2023mononerf, luiten2023dynamic,li2020real,  li2023efficient, wang2024shape, luiten2023dynamic}. With the great success of NeRF in rendering novel-view for static scenes by using implicit representation, many methods~\cite{du2021neural, muller2022instant, cao2023hexplane, noguchi2021neural,xiao2025neural, fridovich2023k} have extended NeRF to model dynamic scenes and have shown a promising visual quality. However, it is challenging for NeRF-based approaches to achieve real-time rendering due to a large number of sampling points and network forward computation for each sampled point alone each light ray. 

Recently, 3DGS~\cite{kerbl20233d} provides an explicit representation with anisotropic 3D Gaussians for the static scene, and its representation and tile-based rasterization allows fast training and real-time rendering. Different approaches~\cite{lin2024gaussian, luiten2023dynamic, yang2024deformable} extend this representation for modelling dynamic scenes. The straightforward approach is to optimize per-frame 3DGS~\cite{luiten2023dynamic}, which can recover the physical trajectories of Gaussian points. Still, it requires multi-view images and large storage memory and cannot be generalized to monocular video inputs. Other approaches~\cite{yang2024deformable, wu20244d} decouple dynamic scenes into a set of 3DGS in canonical space and an implicit neural field (MLPs) for modelling dynamic deformation. However, they usually require computationally expensive forward passes of the neural network, lowering the rendering speed of the original 3DGS, besides, the neural deformation field models deformation independently for each timestamp without considering temporal consistency and motion prior, these implicit representations fail to capture the intricate details of objects in the scene and cannot generate physically plausible motion trajectories. 

The reconstruction of dynamic scenes for both novel-view synthesis and underlying motion from a monocular video is a challenging and under-constrained optimization problem. To tackle it, we assume that the underlying motion of the dynamic foreground shares several non-rigid motion trajectory bases~\cite{akhter2008nonrigid, wang2021neural, li2023dynibar}, and then propose to model a dynamic scene with a motion trajectory field in 3GS representation. The motion trajectory field, modelling dynamic scenes with a compact and regular motion representation, can handle complex object motions precisely and produces high-quality renderings with physically plausible motion awareness~\cite{li2023dynibar, wang2021neural}. We model the dynamic scene learnable motion trajectory basis and lightweight multi-layer perceptron (MLP) for motion coefficient, which allows fast rendering. Sharing global motion basis across Gaussians encourages neighbouring points to have similar motion. The time-invariant motion coefficients and the small number of motion trajectory bases greatly reduce the optimization complexity, which helps to capture the underlying intricate motion pattern. To better optimize the motion trajectory field and reduce the number of Gaussian primitives, we decouple the dynamic scenes into dynamic objects and static backgrounds, so that the motion trajectory field can be modelled compactly without requiring too many 3DGS points, as it has been observed that model entire scenes with a deformation field has the issue of over-consuming GPU memory cased by over-densification. Our representation can provide an easy way to do regularization on motion trajectory and reduce the complexity of this optimization problem. The experimental results show that our work can achieve high-fidelity novel-view synthesis and meanwhile, recover the underlying motion trajectories from a monocular video with real-time rendering. In summary, our main contributions include threefold: 1). we propose to combine motion trajectory field and 3DGS for dynamic scene reconstruction; 2). we design the specialized regularization terms for static/dynamic segmentation and motion trajectory recovery; 3). we have conducted extensive quantitative and qualitative experiments to evaluate the performance of our method.

\vspace{-0.1em}
\section{Related work}
\vspace{-0.4em}
\subsection{Dynamic view synthesis}
View synthesis for dynamic scenes has become a significant research focus in 3D vision, particularly with the development of NeRF~\cite{mildenhall2020nerf}. NeRF-based methods have demonstrated exceptional results in novel view synthesis by implicitly modelling 3D scenes. Extensions~\cite{li2022neural, li2021neural, pumarola2021d, gao2021dynamic, tian2023mononerf, liu2023robust} of NeRF to dynamic scenes have utilized time-conditioned latent codes and explicit deformation fields to capture temporal variations. These approaches aim to model scene motion and appearance changes, though the high computational costs and slow rendering times have limited their practical use. Methods such as hash encoding~\cite{muller2022instant, wang2024masked}, explicit voxel grid~\cite{fang2022fast, gan2023v4d} and feature grid planes~\cite{cao2023hexplane, fridovich2023k, shao2023tensor4d} have accelerated training and improved dynamic scene handling. However, real-time rendering remains challenging for NeRF-based methods due to the computationally expensive volume rendering process. Recently, 3DGS~\cite{kerbl20233d} has emerged as an efficient alternative for dynamic scene reconstruction and view synthesis. Unlike NeRF, 3DGS leverages explicit 3D Gaussian representations combined with differentiable splatting, allowing for real-time rendering without the need for expensive volume rendering. Various methods ~\cite{yang2023real, kratimenos2023dynmf, yang2024deformable, katsumata2023efficient, luiten2023dynamic} have extended 3DGS to dynamic scenes. ~\cite{luiten2023dynamic} pioneered the dynamic 3DGS by iteratively optimizing Gaussian parameters frame-by-frame. ~\cite{wu20244d, yang2024deformable, katsumata2023efficient} employs a deformation field to model Gaussian transformations across time. These approaches effectively reduce memory consumption and improve efficiency, but challenges like motion ambiguities and maintaining temporal consistency persist.

\subsection{Deformation recovery from dynamic scenes}
Recovering deforming 3D shapes from a monocular video is important for scene understanding~\cite{bregler2000recovering}. The 2D point correspondences across temporal frames are usually required for modelling deformation~\cite{akhter2010trajectory, torresani2008nonrigid, kumar2017spatio, zhu2013convolutional, park20103d}. ~\cite{akhter2010trajectory} proposes a trajectory space for the motion of 3D points and assumes that trajectories of points are a composite of a small number of discrete cosine transform (DCT) trajectory bases, and its variants include convolutional trajectory structure~\cite{kumar2017spatio} and trajectory subspace~\cite{kumar2017spatio}. With the success of NeRF, ~\cite{wang2021neural} has integrated trajectory space into NeRF by representing the motion of 3D sampled points with DCT trajectory bases for modelling dynamic scenes. To model long videos, ~\cite{li2023dynibar} introduces learnable motion trajectory basis functions to extend its modelling capabilities. However, these implicit representations still require a long training time and suffer from slow inference. To model the deformation with explicit 3D Gaussian representation, ~\cite{luiten2023dynamic} optimize a set of Gaussian primitives at the first frame and obtain their trajectories across all frames through per-frame optimization, but this approach requires multi-view images to provide a well-constrained optimization process. Gaussian-flow~\cite{lin2024gaussian} models the motion trajectory of Gaussian points in both the time and frequency domains and the time domain. ~\cite{guo2024motion} introduces the optical flow to enhance the motion modelling of dynamic scenes. Shape-of-motion~\cite{wang2024shape} utilizes depth maps and 2D tracks for constraining the motion of Gaussian primitives. However, these methods usually require strong data-driven priors, which could be noisy and degenerate the performance. In this work, we propose to use learnable motion trajectory bases to model the motion of 3DGS, which can not only achieve real-time rendering, but also recover of underlying motion trajectory of 3D Gaussian primitives.

\section{Methodology}

\subsection{Preliminary: 3D points in trajectory space}
\vspace{-0.3em}
All deformable 3D points can be represented into low-rank trajectory space~\cite{akhter2008nonrigid}. To represent the structure at a specific time $t$, we organize the 3D coordinates of the points into a matrix $\boldsymbol{x}^t \in \mathbb{R}^{3 \times P}$, $\boldsymbol{x}^t = \begin{bmatrix} x^t_{1} & x^t_{2} & \cdots & x^t_{P} \end{bmatrix}$,
where ${x}_{ti} \in \mathbb{R}^3$ denotes the 3D coordinates of point $i$ at time $t$. The complete time-varying structure is represented by concatenating these deformable structures into $X = \begin{bmatrix} \boldsymbol{x}^1 \;  \boldsymbol{x}^2 \;  \dots \; \boldsymbol{x}^N \end{bmatrix}^{T}\in \mathbb{R}^{(N \times 3) \times P}$. The row space of this matrix represents the shape space~\cite{xiao2004closed}, while the column space is referred to as the trajectory space. For the shape and motion recovery, the structures are usually effectively represented using $k$ ($k \ll P$) basis shapes~\cite{torresani2008nonrigid}, since the trajectory space holds a dual relationship with the shape space, the trajectory space can be equivalently represented with other $k$ basis trajectory vectors. To express the time-varying structure using trajectory bases, we view the deformation as a set of trajectories $T_i \in \mathbb{R}^{N\times 3}$, i.e. $T_i = \begin{bmatrix} x^1_{i} & x^2_{i} & \cdots & x^N_{i} \end{bmatrix}^T.$

Each trajectory ${T}_i$ can be described as a linear combination of basis trajectories: $T_i = \sum_{j=1}^k a_{ij} {\Theta}_j$, where ${\Theta}_j \in \mathbb{R}^{N \times 3}$ is a basis trajectory, and $a_{ij}$ are the coefficients corresponding to that basis vector. The time-varying structure matrix $X$ can then be factorized into an inverse projection matrix ${\Theta} \in \mathbb{R}^{(N \times 3) \times k}$ and a coefficient matrix $A \in \mathbb{R}^{k \times P}$, i.e. $X = {\Theta} A$, where $A = \begin{bmatrix} \boldsymbol{a}_1 & \boldsymbol{a}_2 & \cdots & \boldsymbol{a}_P \end{bmatrix}$, $\boldsymbol{a}_i = \begin{bmatrix} a_{i1} & a_{i2} & \cdots & a_{ik} \end{bmatrix}^{T}, \quad {\Theta} = \begin{bmatrix} {\Theta}_1 & {\Theta}_2 & \cdots & {\Theta}_k \end{bmatrix}$.

The principal benefit of the trajectory space representation is that a basis can be pre-defined and can compactly approximate most real trajectories. Various bases such as the Hadamard Transform, Discrete Fourier Transform, and Discrete Wavelet Transform~\cite{wang2021neural, li2024generative} can all represent trajectories in an object-independent manner. This prior information will relieve the burden of optimizing the motion of 3D points in a dynamic scene. This paper uses the DCT basis to initialize all basis trajectories, allowing optimization with a good initialization.

\subsection{3D Gaussians in trajectory space}
Given a sequence of images from a dynamic scene with frames ($I_1$, $I_2$,...,$I_N$), and known camera parameters, our method aims to synthesize the novel-view image at any time point to recover the geometric dynamic, such as point trajectory.

\begin{figure*}
    \centering
    \includegraphics[width=1.0\linewidth]{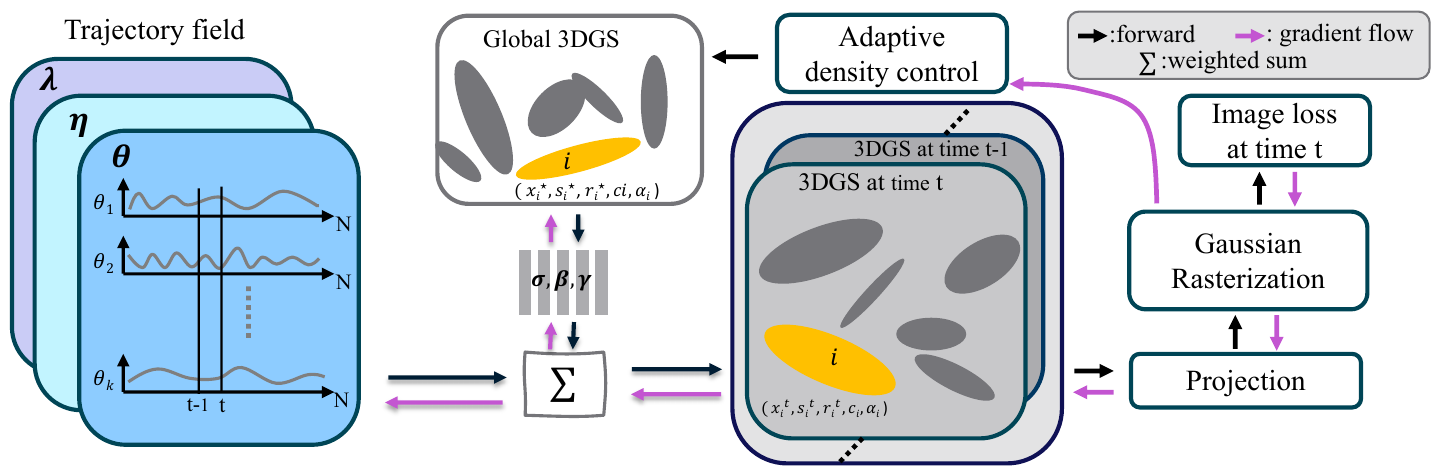}
    \vspace{-0.2em}
    \caption{The pipeline of our framework. 3DGS at time $t$ are transformed from global 3DGS using learnable motion bases ($\theta, \lambda, \eta$) and trajectory coefficients ($\sigma, \beta, \gamma$). Projection, density control, and Gaussian rasterization follow original paper~\cite{kerbl20233d}.}
    \vspace{-0.0em}
    \label{fig:enter-label}
\end{figure*}

3D Guassians~\cite{kerbl20233d} was originally designed to represent static scenes. To extend its capabilities for dynamic scenes, we model the dynamics using a motion trajectory field. We assume a set of compact and moving 3D Gaussian points represents the dynamic scene, where each point follows a trajectory across all time frames. Each 3D Gaussian point has a time-varying covariance matrix but a time-invariant color coefficient (i.e., spherical harmonics). For the time-varying structure, we define a global reference point $x^{\star}_i \in \mathbb{R}^3$ for each trajectory $T_i$, which represents the global position of the point throughout the entire sequence. The global reference point $x^{\star}_i$ can be the initial position or a point in the middle of the trajectory $T_i$, differing from points in canonical space~\cite{pumarola2021d,yang2024deformable,liu2024modgs}. There are several reasons for defining the global reference point $x^{\star}_i$: 1). it helps mitigate global offsets by representing structures relative to global points, leading to more accurate and stable reconstructions; 2). global points can be flexibly selected or learned for different types of motion and deformation; 3). global points are useful for adaptive density control in 3DGS during optimization, helping prune transparent Gaussian points while densifying points in high-frequency areas. By decoupling global points and focusing on relative structures, we achieve a more robust and accurate process for deformation reconstruction.

We model the dynamic scene using relative time-varying structures, as shown in~\cref{fig:enter-label}. The relative position of the $i$th 3D Gaussian at time $t$ is given by $\Delta x_i^t = x_i^t - {x}^{\star}_i $, where $x_i^t$ is the center of the $i$th 3D Gaussian. Therefore, the relative trajectory of the $i$th 3D Gaussian is $\Delta T_i = [\Delta x_i^0, \Delta x_i^1, \dots, \Delta x_i^N]^{T}$, and the relative motion trajectories for all 3D Gaussian points can be represented as $\Delta T = [\Delta T_0, \Delta T_1, \dots, \Delta T_P] \in \mathbb{R}^{(N\times3)\times P}$. The relative trajectory of the $i$th point can be expressed as a linear combination of basis trajectories: $\Delta T_i = \sum_{j=1}^{k} {\sigma}^x_j({x}^{\star}_i)  \theta_j$, where $\theta_j \in \mathbb{R}^{N}$ are the trajectory basis vectors over all time frames, $\theta_j = [\theta_j^0, \theta_j^1, \dots, \theta_j^N]^{T}$, and ${\sigma}^x_j({x}^{\star}_i) \in \mathbb{R}^3$ are the trajectory coefficients modeled with MLPs and shared across all time frames. All global points are detached when inputted into the motion trajectory field, without background propagation. For the $i$th 3D Gaussian point at time $t$, its relative and global positions can be represented as:

\begin{equation}
\centering
\Delta x_i^t  = \sum_{j=1}^{k} {\sigma}_j({x}^{\star}_i) \theta_j^t 
\label{equ:1}
\end{equation}

\begin{equation}
 x_i^t  = {x}^{\star}_i + \sum_{j=1}^{k} {\sigma}_j({x}^{\star}_i) \theta_j^t   
 \label{equ:2}
\end{equation}

We represent scene motion using a motion trajectory field, described through learnable basis functions. For each 3D Gaussian point, we encode its trajectory coefficients with an MLP $\mathcal{F}_{tc}$, as follows:

\begin{equation}
\{\sigma_j(x^{\star}_i)\}_{j=1}^k = \mathcal{F}_{tc}(\mathcal{G}({x}^{\star}_i))
\end{equation}

where $\sigma_j \in \mathbb{R}^3$ are basis coefficients (separate for $x$, $y$, and $z$) and $\mathcal{G}$ represents positional encoding. We choose $k=40$ basis functions. The encoding function $\mathcal{G}$, with linearly increasing frequency, is expressed as:

\begin{equation}
\mathcal{G}({x}^{\star}_i) = \left( \sin(2^k \pi {x}^{\star}_i), \cos(2^k \pi {x}^{\star}_i) \right)_{k=0}^{L-1}
\end{equation}

where $L=12$ for encoding global reference point ${x}^{\star}_i$. This choice is based on the assumption that scene motion tends to occur at low frequencies~\cite{zhang2021consistent}.

The global learnable motion basis, $\{\theta_j^t\}_{j=1}^k$, where $\theta_j^t \in \mathbb{R}$, is introduced to replace the trajectories basis in the original trajectory space~\cite{akhter2008nonrigid}. These bases span every time step $t$ of the input video and are optimized jointly with the MLP. Similarly with~\cite{wang2021neural, li2023dynibar}, we initialize the basis $\{\theta_j\}_{j=1}^k$ using the DCT basis, but fine-tune it during optimization along with other components. This fine-tuning is necessary because a fixed DCT basis often fails to capture the wide range of real-world motions~\cite{li2023dynibar}. The global reference points ${x}^{\star}_i$ are initialized using points from COLMAP~\cite{schoenberger2016sfm}, a standard procedure in 3DGS~\cite{kerbl20233d}.

Using ~\cref{equ:1} and~\cref{equ:2}, we can generate all Gaussian central points across all time frames, ${X} = [\boldsymbol{x}^1, \boldsymbol{x}^2, ...,\boldsymbol{x}^N]^{T} \in \mathbb{R}^{(N\times 3)\times P}$, each row corresponds to all points at a single time frame, while each column represents the trajectory of a point. This motion trajectory field models the dynamic 3D Gaussian points. However, we observed that if each trajectory shares the same covariance across all time frames, using only a time-varying Gaussian centre is insufficient to effectively model dynamic scenes. Since 3DGS includes anisotropic covariance, each Gaussian may exhibit temporally different rotations to capture dynamic geometry and appearance changes. To better model scene deformations, we extend the motion trajectory field (~\cref{equ:1} and~\cref{equ:2}), to also account for time-varying covariance $\Sigma_i = \boldsymbol{r}_i \boldsymbol{s}_i \boldsymbol{s}_i^{T} \boldsymbol{r}_i^{T}$~\cite{kerbl20233d}, and we use $l$ motion-scale bases and $m$ rotation bases, as the following equations:

\begin{equation}
\centering
s^t_i  = {s}^{\star}_i + \sum_{j=1}^{l} {\beta}_j({x}^{\star}_i) \lambda_j^t 
\end{equation}

\begin{equation}
r^t_i  = {r}^{\star}_i + \sum_{j=1}^{m} {\gamma}_j({x}^{\star}_i) \eta_j^t
\label{equ:scale}
\end{equation}

To simplify the model,  we keep opacity and radiance time-invariant, sharing them across all times and global reference points. The geometrical structure and covariance are time-varying, but each trajectory shares the same spherical harmonics coefficients, as the covariance and central points already account for changes in viewing perspectives over time. For the Gaussian $i$ at time $t$, we can obtain its Gaussian primitives \(\{x^t_i, s^t_i, r^t_i, \alpha_i, c_i\}\). By applying rendering equations from~\cite{kerbl20233d}, we can render the image at any time frame.

\subsection{Static and dynamic separation}

\begin{figure}
    \centering
    \includegraphics[width=1.\linewidth]{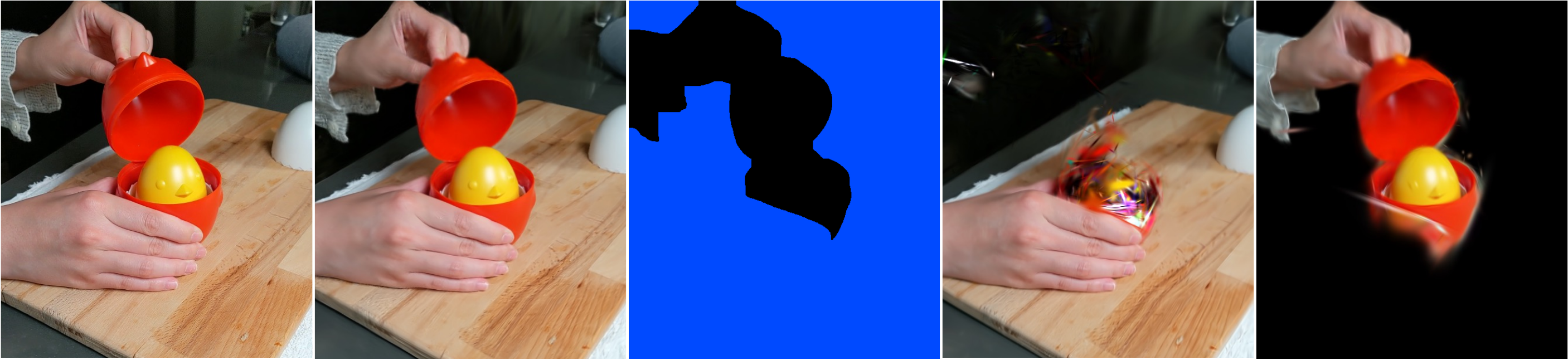}
    \caption{Visualization of static and dynamic Gaussian primitives. The images, from left to right, show the ground truth, rendered image, foreground mask, rendered static scene, and rendered dynamic object. For the foreground mask, black represents the dynamic foreground.}
    \label{fig:mask_img}
\end{figure}

We assign an additional indicator parameter to each 3D Gaussian primitive to represent its states (static or dynamic). This distinction allows us to select dynamic Gaussian points for supervising the motion field. When modeling entire dynamic scenes using an implicit deformation field~\cite{yang2024deformable}, we observed that the static part can mistakenly acquire trajectories (see~\cref{fig:mask_img}). This not only harms novel-view synthesis for the static part but also introduces noisy supervision during training neural motion field. To address this, we assign a continuous parameter \(p \in [0, 1]\) to indicate the probability of a Gaussian primitive belonging to either the static background or dynamic foreground. Ideally, dynamic Gaussians in the foreground should have \(p = 1\), while others in the background should have \(p = 0\). Thus, the parameters of each 3D Gaussian primitive are extended to \(\{x^t_i, s^t_i, r^t_i, \alpha_i, c_i, p_i\}\). We render a static/dynamic segmentation mask ($\mathcal{\hat{M}}$) using volume rendering by replacing the $c_i$ with $p_i$ in rendering equations~\cite{kerbl20233d}. A pseudo-segmentation mask ($\mathcal{M}$) for dynamic objects is generated using segment and tracking anything~\cite{kirillov2023segment, ke2024segment, yang2023track}. The segmentation loss can be obtained as follows:

\begin{equation}
    \mathcal{L}_{m} = ||\mathcal{\hat{M}}  - \mathcal{M}||^2
\end{equation}

When rendering the segmentation map, Gaussian attributes of location, covariance, and opacity are detached without gradient, so that this loss only optimizes its state probability \(p_i\). We set a high threshold (0.8) for selecting the static Gaussian to mitigate the impact of segmentation errors on rendering and motion modelling, as it is observed that the motion field can model static background but not the other way around. The parameter \(p_i\) indicates whether a Gaussian is dynamic. When the model is well-optimized, \(p_i\) should be either \(0\) or \(1\), based on which we design the point-wise loss \(\mathcal{L}_{3mr}\) to force each Gaussian to have only one state. We add this loss at the late stage of the training process.

\begin{equation}
    \mathcal{L}_{3mr} = {\textstyle\footnotesize \frac{1}{k} \sum_{i=1}^{k}} -\left(p_i \log (p_i) + (1 - p_i) \log (1 - p_i)\right)
\end{equation}

\subsection{Motion regularization}
In the motion trajectory field, each point only provides location-dependent coefficients and all points share the same global motion basis which provides a weak prior on a smooth trajectory. To inject physical-based priors into the motion trajectory field, we apply the as rigid as possible loss~\cite{kumar2017monocular, luiten2023dynamic}, $\mathcal{L}^{\text{arap}}$, to achieve the temporal smoothness. We assume that each Gaussian and their neighboring Gaussian points should follow rigid transformation of the coordinate system cross two timesteps. The $\mathcal{L}^{\text{arap}}$ is defined as:
\begin{equation}
\begin{aligned}
\mathcal{L}_{arap} = \frac{1}{k|\mathcal{S}|} \sum_{i \in \mathcal{S}} \sum_{j \in \text{knn}_{i}}  &  w_{i,j} \| ( x_{j}^{t-1} - x_{i}^{t-1}) - \\ & r_{i}^{t-1} (r_{i}^t)^{-1}( x_{j}^t - x_{i}^t) \|_2       
\end{aligned}
\end{equation}

We randomly select $\mathcal{S}$ points to enforce the $\mathcal{L}^{\text{arap}}$, and for each selected point, a set of neighboring Gaussians is selected for it using the k-nearest neighbors (knn), and the loss is down-weighted by an isotropic Gaussian weighting factor:
\[
w_{i,j} = \exp\left( -\rho_w \| x_{j}(t) - x_{i}(t) \|_2^2 \right)
\]

We proposed a spatial smoothness loss to enforce smoothness over neighbouring spatial locations. We apply a perturbing $\epsilon$ on the input global points $x^{\star}$ and encourage the location-dependent attributes at location ${x}^{\star}_i + \epsilon$ to be consistent with ${x}^{\star}_i$. The spatial smoothness term is:

\begin{equation}
\begin{aligned}
    \mathcal{L}_{sp} = & \| {\sigma}_j({x}^{\star}_i) - {\sigma}_j({x}^{\star}_i + \epsilon) \|_2  + \\  
                     & w_{\beta} \| {\beta}_j({x}^{\star}_i) - {\beta}_j({x}^{\star}_i + \epsilon) \|_2  +  \\
                   & w_{\gamma} \| {\gamma}_j({x}^{\star}_i) - {\gamma}_j({x}^{\star}_i + \epsilon) \|_2    
\end{aligned}
\end{equation}

\noindent where the perturbing value's magnitude is adaptively set according to the scale of the scene, i.e., $\epsilon = 0.001*scale$, and $w_{\beta}$ and $w_{\gamma}$ are corresponding weighting factor for rotation and scale.

All these motion regularisers are applied only on dynamic Gaussian points using the foreground/background mask. We observe that dynamic Gaussian point tends to have a larger gradient flow than static points, since dynamic region has a large photometric loss and is also constrained with motion regularisers. If we apply the similar density control mechanism to both points, this can cause over-dense dynamic point; therefore, we set a large gradient threshold for densifying the dynamic points.

When the model is well optimized, we can render the image with the learned Gaussian primitives \(\{\mathbf{x}^*_i, \mathbf{s}^*_i, \mathbf{r}^*_i, \alpha_i, c_i, p_i\}\) with trajectory motion field only applied on dynamic Gaussian points. With the rendered images, we can calculate the Photometric loss $\mathcal{L}_{pho}$, which is the same as~\cite{kerbl20233d}. Therefore, our final loss to optimize the model is as follows: 
\begin{equation}
\mathcal{L} =  \mathcal{L}_{pho} + \mathcal{L}_{m} + \mathcal{L}_{3mr} +  \lambda_a\mathcal{L}_{arap}  + \lambda_s\mathcal{L}_{sp}   
\end{equation}

\section{Experiments}

\subsection{Experimental setup}
\subsubsection{Dataset:}  We conduct various experimental evalution on two dataset: D-NeRF~\cite{pumarola2021d} and HyperNeRF~\cite{park2021hypernerf}, and both are monocular video dataset. D-NeRF is a synthetic dataset, including 8 sets of dynamic scenes, and each scene features complex motion such as articulated objects and human actions. Every image is 800$\times$800 with 100 to 200 images per scene. Most images only contain the dynamic object without static background, therefore we do not apply the mask loss, i.e. $\mathcal{L}_{3mr}$ and $\mathcal{L}_{m}$, for this dataset. HyperNeRF is a real-world dataset that includes a monocular video on both real rigid and non-rigid deformable objects. We create a mask for each object using track anything with one click~\cite{yang2023track}. All images are down-sampled to 540$\times$960 for a fair comparison with other baselines.

\subsubsection{Implementation details:} The entire training iterations are 50k, and we firstly only train the global Gaussian points up to 5k iterations without incorporating trajectory fields for stable training, then we start to optimize the trajectory motion field jointly until the end and stop the densification until 25k iteration. The Adam optimizer~\cite{KingBa15} is used to optimize our model with different learning rates for each module, and the step learning rate (StepLR) scheduler in PyTorch~\cite{paszke2019pytorch} is used to adjust the learning rate. The learning rate for optimizing the 3DGS is the same as the original paper~\cite{kerbl20233d}. For optimizing the trajectory motion field, we use a learning rate of $1e-3$ for the coefficient MLP model with a decay factor of 0.5 for every 15k iterations in StepLR and use the learning rate $5e-4$ for trajectory basis with the same decay factors. The gradient thresholds for densification are $4e-4$ and $8e-4$ for static and dynamic points respectively. $\lambda_a$ and $\lambda_s$ are set to 0.3 and 0.6 respectively. both of $w_{\beta}$ and $w_{\gamma}$ are 0.5. The 3DGS are initialized randomly for D-NeRF~\cite{pumarola2021d} and The initialization for HyperNeRF~\cite{park2021hypernerf} is from Structure-from-motion points derived from COLMAP~\cite{schoenberger2016sfm}.


\subsection{Comparison results}

\begin{table*}[h]
\centering
\caption{Quantitative comparison of our method and baselines evaluated on the synthetic dataset~\cite{pumarola2021d}. We report three metrics: peak signal-to-noise ratio (PSNR), structural similarity index (SSIM), and learned perceptual image patch similarity (LPIPS) with VGG model~\cite{zhang2018perceptual}, for eight scenes with full image resolution 800$\times$800. The color coding in the table highlights the \colorbox{best}{best}, \colorbox{second}{second}-best, and \colorbox{third}{third}-best performance. The results of baselines are gathered from the paper on the corresponding methods.}
\begin{adjustbox}{width=1.0\textwidth}
\begin{tabular}{l|ccc|ccc|ccc|ccc}
\toprule
\multirow{2}{*}{Method} & \multicolumn{3}{c|}{Hell Warrior}  & \multicolumn{3}{c|}{Mutant} & \multicolumn{3}{c|}{Hook} & \multicolumn{3}{c}{Bouncing Balls} \\
\cmidrule(lr){2-4} \cmidrule(lr){5-7} \cmidrule(lr){8-10} \cmidrule(lr){11-13}
  & PSNR↑ & SSIM↑ & LPIPS↓ & PSNR↑ & SSIM↑ & LPIPS↓ & PSNR↑ & SSIM↑ & LPIPS↓ & PSNR↑ & SSIM↑ & LPIPS↓ \\
\midrule
3D-GS    & 29.89 & 0.9155 & 0.1056 & 24.53 & 0.9336 & 0.0580 & 21.71 & 0.8876 & 0.1034 & 23.20 & 0.9591 & 0.0600 \\
D-NeRF   & 24.06 & 0.9440 & 0.0707 & 30.31 & 0.9672 & 0.0392 & 29.02 & 0.9595 & \cellcolor{third} 0.0546 & 38.17 & 0.9891 & 0.0323 \\
TiNeuVox & 27.10 &\cellcolor{third} 0.9638 & 0.0768 & 31.87 & 0.9607 & 0.0474 & \cellcolor{third}30.61 & \cellcolor{third} 0.9599 & 0.0592 & \cellcolor{second} 40.23 & 0.9926 & 0.0416 \\
Tensor4D & \cellcolor{third}31.26 & 0.9254 & \cellcolor{third}0.0735 & 29.11 & 0.9451 & 0.0640 & 28.63 & 0.9433 & 0.0636 & 24.47 & 0.9622 & 0.0437 \\
K-Planes & 24.58 & 0.9520 & 0.0824 & \cellcolor{third}32.50 & \cellcolor{third}0.9713 & \cellcolor{third}0.0362 & 28.12 & 0.9489 & 0.0662 & 40.05 & \cellcolor{second}0.9934 & \cellcolor{third}0.0322 \\
D-3DGS    & \cellcolor{second}41.54 & \cellcolor{second}0.9873 & \cellcolor{second}0.0234 & \cellcolor{second}42.63 & \cellcolor{second}0.9951 & \cellcolor{second}0.0052 &\cellcolor{second} 37.42 &\cellcolor{second} 0.9867 & \cellcolor{second}0.0144 & \cellcolor{best} 41.01 & \cellcolor{best}0.9953 & \cellcolor{best}0.0093 \\
Ours & \cellcolor{best}41.67& \cellcolor{best}0.9877 & \cellcolor{best}0.02361 & \cellcolor{best}43.27 & \cellcolor{best}0.9962 & \cellcolor{best} 0.0043 & \cellcolor{best}38.68& \cellcolor{best}0.989& \cellcolor{best}0.0115 & \cellcolor{third}40.15 & \cellcolor{third} 0.9929 & \cellcolor{second} 0.0297 \\ 
\midrule
& \multicolumn{3}{c|}{Lego} & \multicolumn{3}{c|}{T-Rex} & \multicolumn{3}{c|}{Stand Up} & \multicolumn{3}{c}{Jumping Jacks} \\
\cmidrule(lr){2-4} \cmidrule(lr){5-7} \cmidrule(lr){8-10} \cmidrule(lr){11-13}
Method & PSNR↑ & SSIM↑ & LPIPS↓ & PSNR↑ & SSIM↑ & LPIPS↓ & PSNR↑ & SSIM↑ & LPIPS↓ & PSNR↑ & SSIM↑ & LPIPS↓ \\
\midrule
3D-GS    & 22.10 & 0.9384 & 0.0607 & 21.93 & 0.9539 & 0.0487 & 21.91 & 0.9301 & 0.0785 & 20.64 & 0.9297 & 0.0828 \\
D-NeRF   & 25.56 & 0.9363 & 0.0821 & 30.61 & 0.9671 & 0.0535 & 33.13 & 0.9781 & 0.0355 & 32.70 & \cellcolor{third}0.9779 & \cellcolor{third}0.0388 \\
TiNeuVox & 26.64 & 0.9258 & 0.0877 & \cellcolor{third}31.25 & 0.9666 & 0.0478 & \cellcolor{third} 34.61 &\cellcolor{third} 0.9797 & 0.0326 & \cellcolor{third}33.49 & 0.9771 & 0.0408 \\
Tensor4D & 23.24 & 0.9183 & 0.0721 & 23.86 & 0.9351 & 0.0544 & 30.56 & 0.9581 & 0.0363 & 24.20 & 0.9253 & 0.0667 \\
K-Planes & \cellcolor{third}28.91 & \cellcolor{third}0.9695 & \cellcolor{third}0.0331 & 30.43 & \cellcolor{third}0.9737 & \cellcolor{third}0.0310 & 33.10 & 0.9793 & \cellcolor{third} 0.0310 & 31.11 & 0.9708 & 0.0468 \\
D-3DGS   & \cellcolor{best}33.07 & \cellcolor{best}0.9794 &\cellcolor{best} 0.0183 & \cellcolor{second}38.10 & \cellcolor{second}0.9933 & \cellcolor{second} 0.0098 & \cellcolor{second} 44.62 & \cellcolor{second}0.9951 & \cellcolor{second} 0.0063 & \cellcolor{second} 37.72 & \cellcolor{second}0.9897 & \cellcolor{second}0.0126 \\
Ours & \cellcolor{second}30.48 &\cellcolor{second}0.9703 & \cellcolor{second}0.0284 &\cellcolor{best} 38.71 &\cellcolor{best} 0.9939 &\cellcolor{best} 0.0089 &\cellcolor{best} 46.28 & \cellcolor{best}0.9976 & \cellcolor{best} 0.0046 &\cellcolor{best} 40.01 & \cellcolor{best} 0.9930 & \cellcolor{best} 0.0086 \\
\bottomrule
\end{tabular}
\end{adjustbox}
\label{tab:quantitative_syn}
\end{table*}


\subsubsection{Results on synthetic dataset} 
We compared our methods with all relevant baselines with the monocular synthetic dataset from D-NeRF~\cite{pumarola2021d}. We choose 3DGS~\cite{kerbl20233d}, D-NeRF~\cite{pumarola2021d}, TiNeuVox~\cite{fang2022fast}, Tensor4D~\cite{shao2023tensor4d}, K-Planes~\cite{fridovich2023k}, and D-3DGS~\cite{yang2024deformable} as the baselines for this dataset, because we can find the reported performance in their paper for a fair comparison. In this synthesis dataset, static backgrounds are automatically removed and we don't apply the mask loss when optimizing the model. The proposed method (“Ours” in the table) outperforms other baseline methods in most of the scenes, demonstrating superior performance in terms of PSNR, SSIM, and LPIPS, particularly excelling in the Bouncing Balls, T-Rex, Stand Up, and Jumping Jacks scenes. The strong results suggest that it is well-suited for high-quality, perceptually accurate image reconstruction tasks, especially in dynamic synthetic scenes. D-3DGS is a close competitor, showcasing its robustness in certain scenarios, while K-Planes and TiNeuVox also offer competitive alternatives. We can also tell that D-3DGS is another strong performer, showcasing its robustness in certain scenarios. 

\begin{figure*}[ht!]
\centering
\includegraphics[width=1.0\linewidth]{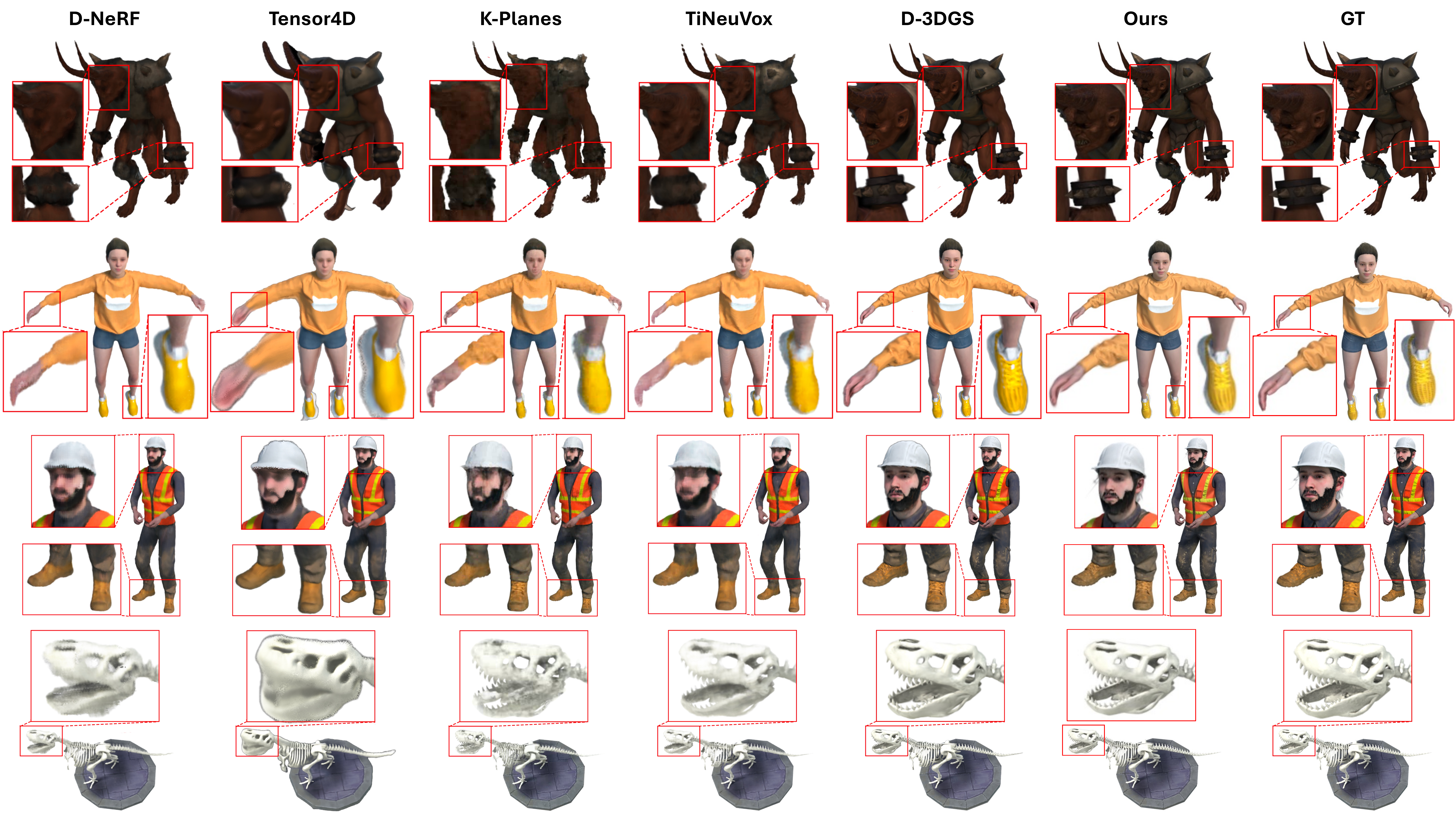}
\caption{Qualitative results of baselines and our method against the ground truth (GT) on a synthetic dataset. The visualization highlights the ability of each method to reconstruct geometrical poses and fine details across different scenes. We visualize the four scenes: HellWarrior, Jumping Jacks, Stand Up, and T-Rex from top to bottom.}
\label{fig:synthesis_img_qualitatives}    
\end{figure*}

\begin{figure}
    \hspace*{-1cm}
    \includegraphics[width=1.1\linewidth]{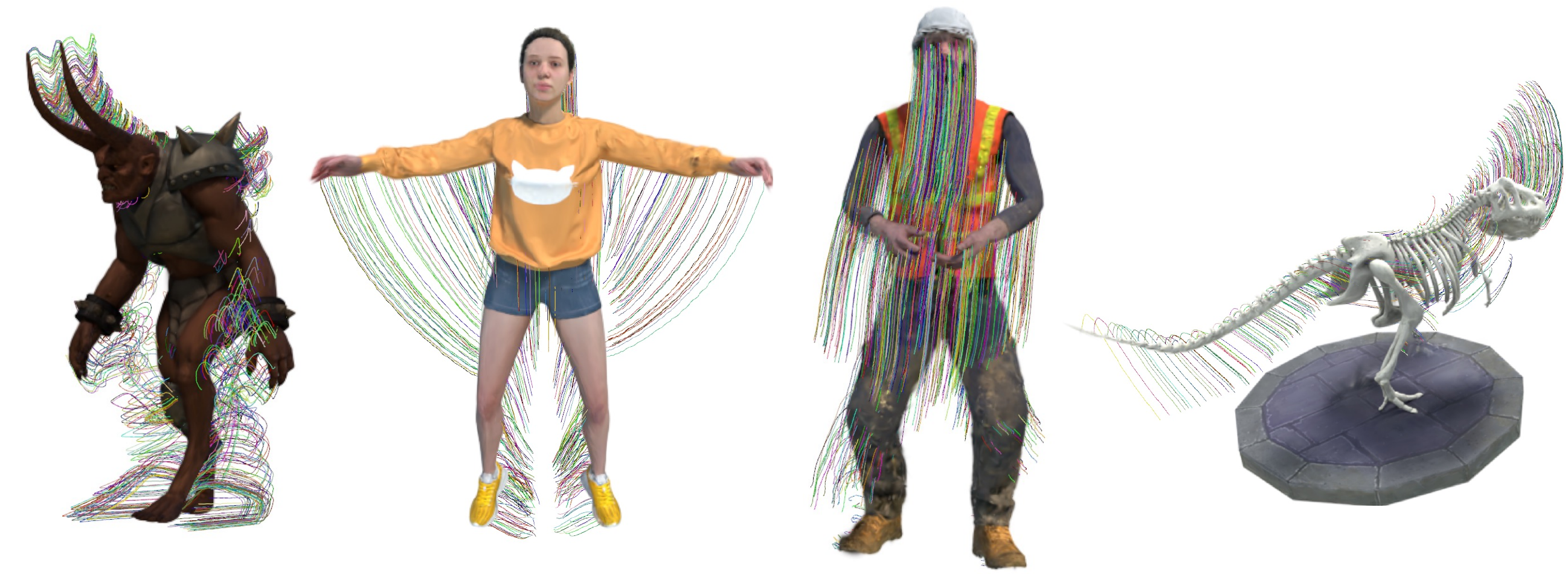}
    \vspace{-0.4em}
    \caption{Visualization of motion trajectory on D-NeRF. }
    \vspace{-0.8em}
    \label{fig:trajectory_synthesis_dataset}
\end{figure}

The qualitative comparison results are shown in ~\cref{fig:synthesis_img_qualitatives}, from which we can tell that our proposed method closely matches the ground truth, preserving finer details and better geometric consistency. While D-3DGS also perform well, capturing many key features, it still falls short in some areas of fine detail preservation. D-NeRF, Tensor4D, and K-Planes struggle with blurring and less accurate geometry, especially noticeable in the finer areas. For instance, our proposed method can model the relative position between the knee- and spiked-wristband on the arm much more accurately than D-3DGS, Tensor4D, and K-planes for the scene of the HellWarrior, meanwhile, our rendering performance is better than TiNeuVox and D-NeRF given the similar geometrical modelling, for the Stand Up, we can tell that our method can preserve the details of hair while other methods failed to model this. The visualization of motion trajectory is visualized in~\cref{fig:trajectory_synthesis_dataset}, in which the motion trajectory can accurately illustrate how objects move. The proposed method stands out by achieving high fidelity and being effective in modelling moving 3D geometry.

\subsubsection{Results on real-world dataset}
\begin{figure}
    \centering
    \includegraphics[width=1.\linewidth]{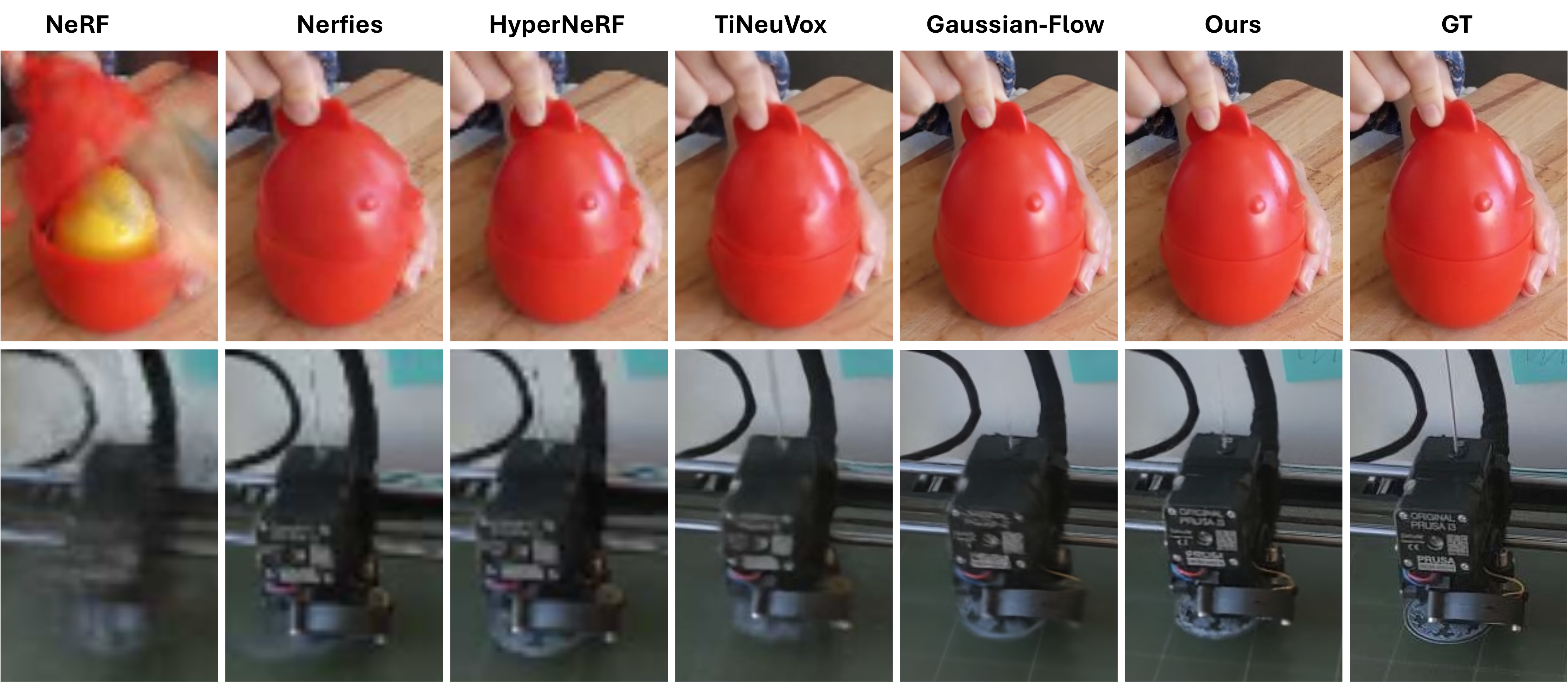}
    \caption{Qualitative comparison. Each row shows results from six methods: NeRF, Nerfies, HyperNeRF, TiNeuVox, Gaussian-Flow, and our method, along with the ground truth (GT).}
    \label{fig:real_img_qualitative}
\end{figure}

We evaluate our method on the real-world dataset HyperNeRF~\cite{park2021hypernerf} and do the comparison with the previous the state-of-the-art: NeRF~\cite{mildenhall2020nerf}, Nerfies, HyperNeRF~\cite{park2021hypernerf}, TiNeuVox~\cite{fang2022fast}, and Gaussian-flow~\cite{lin2024gaussian}. The experimental results, shown in~\cref{tab:real_comparisons}, demonstrate that the proposed method achieves the best performance for novel view synthesis, outperforming other approaches across multiple scenes. It records the highest mean scores in PSNR and SSIM, particularly excelling in the Broom, 3D Printer, and Chicken scenes. While Gaussian-flow~\cite{lin2024gaussian} shows strong performance, particularly in the Peel Banana scene with the highest SSIM, it ranks second overall. TiNeuVox consistently performs well, securing third place in most cases. Comparatively, methods like NeRF, Nerfies, and HyperNeRF show lower performance. These results indicate the effectiveness of the proposed method for high-quality image reconstruction in real-world scenarios. The qualitative comparison for novel view synthesis on two scenes is represented in~\cref{fig:real_img_qualitative}. Our proposed method stands out by producing the most accurate and sharp reconstructions, with sharp details and minimal artifacts, that closely match the ground truth. Other methods, particularly NeRF and Nerfies, exhibit significant blurring and distortions, especially in regions with motion. HyperNeRF, TiNeuVox, and Gaussian-Flow show improvements but still lag behind the visual quality achieved by the proposed method. The motion trajectory is visualized in the rightmost image of~\cref{fig:motion_regularizaiton}.

\begin{table*}[h]
    \centering
    \caption{Per-scene quantitative comparisons on HyperNeRF~\cite{park2021hypernerf} dataset. Results are collected from the corresponding papers. Our method can achieve the best quality for novel view synthesis in this real-world dataset.}
    \label{tab:real_comparisons}
    \begin{adjustbox}{width=.9\textwidth}
    \vspace{-0.4em}
    \begin{tabular}{l|cc|cc|cc|cc|cc}
        \toprule
        \multirow{2}{*}{Method} & \multicolumn{2}{c|}{Broom} & \multicolumn{2}{c|}{3D Printer} & \multicolumn{2}{c|}{Chicken} & \multicolumn{2}{c|}{Peel Banana} & \multicolumn{2}{c}{Mean} \\
\cmidrule(lr){2-3} \cmidrule(lr){4-5} \cmidrule(lr){6-7} \cmidrule(lr){8-9} \cmidrule(lr){10-11}
 & PSNR↑ & SSIM↑ & PSNR↑ & SSIM↑ & PSNR↑ & SSIM↑ &PSNR↑ & SSIM↑ &PSNR↑ & SSIM↑ \\
\midrule
        NeRF  &  19.9 & 0.653 & 20.7 & 0.780 & 19.9 & 0.777 & 20.0 & 0.769 & 20.1 & 0.745 \\
        Nerfies  & 19.2 & 0.567 & 20.6 & 0.830 & 26.7 & 0.943 & 22.4 & 0.872 & 22.2 & 0.803 \\
        HyperNeRF & 19.3 & 0.591 & 20.0 & 0.821 & 26.9 & \cellcolor{third}0.948 & 23.3 & \cellcolor{second}0.896 & 22.4 & 0.814 \\
        TiNeuVox  & \cellcolor{third}21.5 & \cellcolor{third}0.686 & \cellcolor{third}22.8 & \cellcolor{third}0.841 & \cellcolor{third}28.3 & \cellcolor{second}0.947 & \cellcolor{third}24.4 & 0.873 & \cellcolor{third}24.3 &\cellcolor{third} 0.837 \\
        Gaussian-flow  & \cellcolor{second}22.8 & \cellcolor{second}0.709 & \cellcolor{second}25.0 & \cellcolor{second}0.877 & \cellcolor{second}30.4 & 0.945 &  \cellcolor{second} 27.0 &  \cellcolor{best}0.917 &  \cellcolor{second} 26.3 & \cellcolor{second} 0.862 \\
        Ours & \cellcolor{best} 23.4 & \cellcolor{best} 0.745 & \cellcolor{best} 26.7 &\cellcolor{best} 0.896 & \cellcolor{best}31.2 & \cellcolor{best}0.951 & \cellcolor{second}26.2 & \cellcolor{third}0.893 & \cellcolor{best}26.9 & \cellcolor{best}0.871 \\
        \bottomrule
    \end{tabular}
    \end{adjustbox}
    \vspace{-0.8em}
\end{table*}

\subsection{Ablation study}

\subsubsection{static and dynamic separation}

\begin{figure}
    \centering
    \includegraphics[width=1.\linewidth]{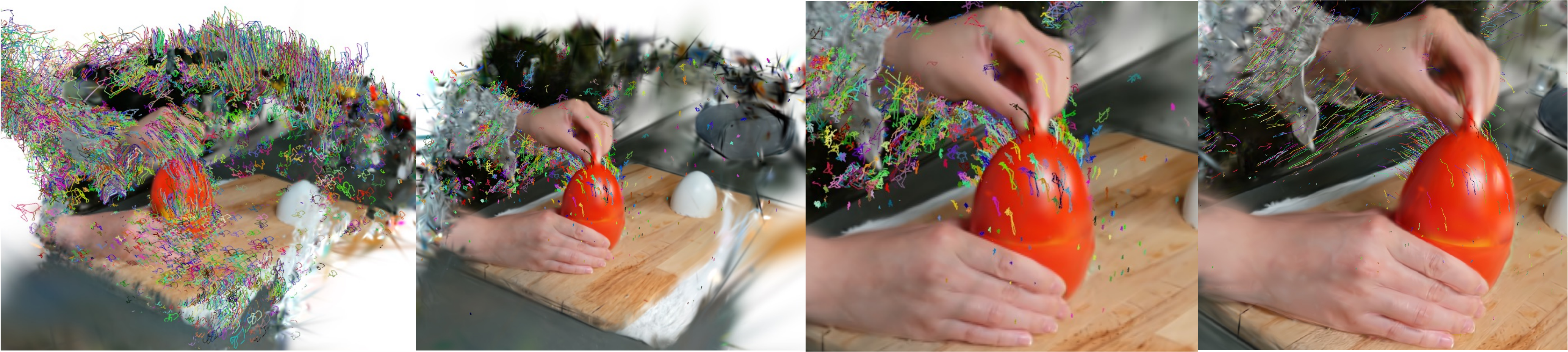}
    \caption{Visualization of motion trajectory in the HyperNeRF dataset. The leftmost image shows the rendered result without static and dynamic separation, the middle two images include static and dynamic separation (one zoomed in), and the rightmost image is rendered from the model with motion regularization.}
    \label{fig:motion_regularizaiton}
\end{figure}

Foreground masks are used to explicitly separate static and dynamic Gaussian primitives. Without separating them, most of the static background will start to move. We can find this from the leftmost image in~\cref{fig:motion_regularizaiton}, from which we can tell that most of the static background points are with trajectory, which can not only degrade the synthesis quality of static background but also cause over-densification of Gaussian points and slow rendering speed. This ablation study evaluates the impact of mask segmentation and its regularization on image quality, speed, and resource efficiency, shown in~\cref{tab:my_label}. We find that introducing mask segmentation ($\mathcal{L}_{m}$) leads to a noticeable improvement, with PSNR increasing to 26.1, SSIM to 0.853, and the speed improving to 22 FPS. Further refining the model by combining mask segmentation with an additional refinement technique ($\mathcal{L}_{m}$ + $\mathcal{L}_{3mr}$) results in the best performance and the highest speed of 26 FPS, using only 480k primitives. The results demonstrate the effectiveness of our mask segmentation module.

\begin{table}[]
    \centering
    \begin{tabular}{c|c c|c c}
    \toprule
          Seg  & PSNR↑ & SSIM↑ & FPS↑ & Num(k) ↓ \\ \midrule
          N/A         & 25.4  & 0.847 & 13 & 740 \\ 
        $\mathcal{L}_{m}$  & 26.1 & 0.853 & 22 & 510\\  
       $\mathcal{L}_{m}$ + $\mathcal{L}_{3mr}$ & 26.9 & 0.871 & 26 & 480 \\
    \bottomrule
    \end{tabular}
    \caption{Ablation study of mask segmentation. Num, the number of Gaussian primitives, is multiplied by $10^3$.}
    \label{tab:my_label}
\end{table}

\subsubsection{Motion Regularization}
~\cref{tab:motion_regularization} and ~\cref{fig:motion_regularizaiton} present the results of an ablation study on motion regularization, comparing different combinations of regularization techniques and their effects on image quality. The baseline `N/A' without any regularization achieves the highest performance, but its trajectory is not smooth. When the as-rigid-as-possible regularization ($\mathcal{L}_{arap}$) is applied, both metrics decrease slightly. Similarly, using spatial smoothness regularization ($\mathcal{L}_{sp}$) shows a minor reduction compared to the baseline. Combining both regularization techniques leads to a decreased performance but with the smoothest trajectory. This suggests that although motion regularization slightly reduces the overall image quality metrics, it contributes positively to motion consistency (as shown in the rightmost image in ~\cref{fig:motion_regularizaiton}, which is critical for certain robotics applications. Noteworthy, our method can still achieve state-of-the-art rendering performance with motion regularization.


\begin{table}[]
    \centering
    \begin{tabular}{c|c c c c}
    \toprule
           & N/A & $\mathcal{L}_{arap}$ & $\mathcal{L}_{sp}$ & $\mathcal{L}_{arap}$ + $\mathcal{L}_{sp}$ \\ \midrule
         PSNR↑ & 27.4  & 27.0 &  27.2  & 26.9 \\
         SSIM↑ & 0.879 & 0.873 & 0.875 & 0.871 \\
    \bottomrule
    \end{tabular}
    \caption{Ablation studies of motion regularization.}
    \label{tab:motion_regularization}
\end{table}



\section{Conclusion}
In this paper, we proposed a novel method for reconstructing dynamic scenes and recovering motion trajectories from monocular video input by combining 3DGS with a motion trajectory field. Our approach effectively handles complex non-rigid motions while achieving real-time rendering performance. The decoupling of dynamic and static components allows for efficient representation and reduced GPU memory consumption. Through extensive quantitative and qualitative evaluations, we demonstrated that our method can produce high-quality novel-view synthesis and physically plausible motion trajectories. The proposed framework provides a significant step forward in the real-time rendering of dynamic scenes and can be a valuable tool for various applications such as virtual reality and robotic manipulation. Future work may explore further optimizations in motion trajectory representation and extend the method to more diverse dynamic scene scenarios.
{
    \small
    \bibliographystyle{ieeenat_fullname}
    \bibliography{main}
}
\end{document}